\documentclass[nohyperref]{article}

\usepackage[accepted]{icml2023}
\usepackage[textsize=tiny]{todonotes}

\icmltitlerunning{Identifying and Disentangling Spurious Features in Pretrained Image Representations}

\usepackage{microtype}

\usepackage{natbib}

\usepackage{array}
\usepackage{amsmath, amssymb, amsthm, bm, bbm, amsfonts, mathalfa}
\usepackage{mathtools}
\usepackage[mathscr]{eucal}
\usepackage{dsfont}

\usepackage[export]{adjustbox} 
\usepackage{tikz}
\usepackage{pgfplots}
\usepackage{graphicx}
\usepackage{subcaption}

\usepackage{algorithm, algorithmic}
\usepackage{listings}

\usepackage{booktabs}
\usepackage{colortbl}
\usepackage{multirow}
\newcolumntype{M}[1]{>{\centering\arraybackslash}m{#1}}
\newcolumntype{L}[1]{>{\raggedright\arraybackslash}m{#1}}

\usepackage{appendix}
\usepackage{spverbatim}
\usepackage{csquotes}
\usepackage{efbox}
\usepackage{enumitem}
\usepackage[shortcuts]{extdash} 
\usepackage[symbol]{footmisc}
\usepackage{hyphenat}
\usepackage{ifthen}
\usepackage{ragged2e}
\usepackage{url}

\usepackage[
  breaklinks    = true,
  colorlinks    = true,
  hypertexnames = false,
  pdfpagelabels = false,
  citecolor     = {blue!80!black},
  linkcolor     = {blue!80!black},
  urlcolor      = {blue!80!black},
]{hyperref} 

\usepackage[capitalize,nameinlink,noabbrev]{cleveref}
\crefformat{equation}{Eq.~(#2#1#3)}
\crefmultiformat{equation}{(#2#1#3)}%
{ and~(#2#1#3)}{, (#2#1#3)}{ and~(#2#1#3)}
\crefformat{theorem}{Theorem~#2#1#3}
\crefformat{proposition}{Proposition~#2#1#3}
\crefformat{lemma}{Lemma~#2#1#3}
\crefformat{remark}{Remark~#2#1#3}
\crefformat{section}{Section~#2#1#3}
\crefformat{assumption}{Assumption~#2#1#3}
\crefformat{example}{Example~#2#1#3}
\crefformat{figure}{Figure~#2#1#3}
\crefformat{algorithm}{Algorithm~#2#1#3}
\crefrangeformat{section}{Section~#3#1#4--#5#2#6}
\crefformat{assumption}{Assumption~#2#1#3}
\crefrangeformat{equation}{~(#3#1#4--#5#2#6)}
\crefrangeformat{figure}{Figure~#3#1#4--#5#2#6}
\crefrangeformat{assumption}{Assumptions~(#3#1#4--#5#2#6)}
\crefformat{appendix}{Appendix~#2#1#3}

\theoremstyle{plain}

\theoremstyle{definition}

\theoremstyle{remark}

\newcommand{\cbr}[1]{\left\{#1\right\}}

\newcommand{\mathset}[1]{\cbr{#1}}  

\newcommand{\best}[1]{\cellcolor{gray!25}{#1}}

\def \bR {\mathbb{R}}

\newcommand{\waterbirds}{\texttt{Waterbirds}}
\newcommand{\captum}{\texttt{Captum}}
\newcommand{\detic}{\texttt{Detic}}

\begin{document}

\twocolumn[
\icmltitle{Identifying and Disentangling Spurious Features in Pretrained Image Representations}

\icmlsetsymbol{equal}{*}

\begin{icmlauthorlist}
\icmlauthor{Rafayel Darbinyan}{ynn,ysu}
\icmlauthor{Hrayr Harutyunyan}{usc}
\icmlauthor{Aram H. Markosyan}{meta}
\icmlauthor{Hrant Khachatrian}{ynn,ysu}
\end{icmlauthorlist}

\icmlaffiliation{ynn}{YerevaNN}
\icmlaffiliation{usc}{University of Southern California}
\icmlaffiliation{meta}{Meta AI Research}
\icmlaffiliation{ysu}{Yerevan State University}

\icmlcorrespondingauthor{Hrayr Harutyunyan}{hrayrhar@usc.edu}

\vskip 0.3in
]

\printAffiliationsAndNotice{}

\begin{abstract}
Neural networks employ spurious correlations in their predictions, resulting in decreased performance when these correlations do not hold.
Recent works suggest fixing pretrained representations and training a classification head that does not use spurious features.
We investigate how spurious features are represented in pretrained representations and explore strategies for removing information about spurious features.
Considering the \waterbirds{} dataset and a few pretrained representations, we find that even with full knowledge of spurious features, their removal is not straightforward due to entangled representation.
To address this, we propose a linear autoencoder training method to separate the representation into core, spurious, and other features.
We propose two effective spurious feature removal approaches that are applied to the encoding and significantly improve classification performance measured by worst group accuracy.
\end{abstract}

\section{Introduction}
In many classification datasets, some features are predictive of the label but are not causally related.
It is often said that these features are \emph{spuriously correlated} with the label, as their correlation might not hold for data collected in another environment.
For example, suppose we collect typical images of cows and camels and form a binary classification task. In that case, we will find that the background is correlated with the label, as cows are often photographed in barns or green pastures, while camels are often photographed in deserts~\citep{beery2018recognition}.
However, this correlation will be \emph{spurious} as the background information is not causally related to the label, and we can easily make another dataset of cows and camels in which this correlation does not hold.

It is well-established that neural networks are susceptible to spurious correlations~\citep{torralba2011unbiased,ribeiro2016should,gururangan-etal-2018-annotation,zech2018variable,mccoy-etal-2019-right,geirhos2018imagenettrained,geirhos2020shortcut,xiao2021noise}.
In such cases, neural networks learn representations that capture spurious features and make predictions that employ them.
Many approaches have been proposed for learning representations that do not capture spurious features~\citep{muandet2013domain,sun2016deep,ganin2016domain,wang2018learning,wang2019learning,li2018deep,arjovsky2019invariant,zhao2020domain,lu2022invariant}.
Some methods are tailored against specific spurious correlations (e.g., texture); some require specifying a categorical spurious feature, while others require data collected from multiple labeled environments.
Nevertheless, to our best knowledge, none of such representation learning methods consistently outperform standard empirical risk minimization~\citep{gulrajani2021in,koh2021wilds}.
This is partly because spurious features are often easier to learn and get learned early in training~\citep{shah2020pitfalls,nam2020learning,hermann2020shapes,pezeshki2021gradient}.

Besides the unsatisfactory results, the approach mentioned above also goes against one of the main techniques of deep learning -- using pretrained representations instead of learning from scratch.
Recently, a few works indicated a large potential in fixing pretrained representations and focusing on training a linear classifier on top of it that does not rely on spurious correlations.
In particular, \citet{galstyan2022failure} find that a significant contribution to the out-of-domain generalization error comes from the classification head and call for designing better methods of training the classification head.
\citet{menon2021overparameterisation} propose to retrain the classification head on training data with down-sampled majority groups.
\citet{kirichenko2023last,izmailov2022feature}; and \citet{shi2023robust} find that after training on data with spurious correlations, keeping the representations fixed and retraining the classification head on small unbiased data gives state-of-the-art results.
When no information about spurious features is available, \citet{mehta2022you} show that one can still get good results by using embeddings from a \emph{large} pretrained vision model.
Interestingly, representations learned by a vision transformer~\citep{dosovitskiy2021an} seem to lead to more robust classification heads~\citep{ghosal2022vision}.
Overall, these findings indicate that more research is needed to understand better how spurious features are represented and design better methods of training classification heads on representations that capture spurious features.

We consider the \waterbirds{} dataset~\citep{Sagawa*2020Distributionally}, which is landbird vs waterbird image classification task where the background is spuriously correlated with the label.
Namely, most landbird images have land in their background, while most waterbird images have water in their background.
We consider fixed pretrained representations learned through  supervised or self-supervised learning.
We investigate whether one can remove the spurious features from the representations in two settings.
In the former (and more prevalent setting), one has access to the value of the binary spurious feature.
In the latter, we also have access to per-example image masks indicating which parts of images correspond the spurious feature.

Interestingly, even with full knowledge of the spurious feature, it is not straightforward to remove it.
While we find that representations are axis-aligned to a certain degree, the extent of alignment is not enough to remove spurious features by removing individual representation coordinates.
Since both the spurious feature and the label can be predicted well from the representation with a linear layer, we hypothesize that the entanglement of core and spurious features is linear and can be reversed with a linear transformation.
For this we propose a linear autoencoder to split the representation into three parts corresponding to the class label, the spurious feature, and other features not related to the former two but required for reconstruction.
Importantly, in contrast to existing approaches, we do not enforce independence of the first and second parts on a biased training set.
Instead, we enforce independence on an upsampled variant of the training set.

We find that a linear classifier trained on the core features of the encoding performs better than the standard approach but does not reach the performance of a classifier trained on an unbiased set.
We demonstrate that this gap can be closed by performing additional feature selection within the core features.

\section{Experimental Design}
\textbf{Waterbirds dataset.}
We consider the \waterbirds{} dataset~\cite{group_dro},  which is a benchmark dataset designed to measure the effect of debiasing spurious correlations. The dataset consists of bird photographs from the \texttt{CUB}~\cite{cub_dataset} dataset combined with image backgrounds from the \texttt{Places}~\cite{places_dataset} dataset. It has 4,795 training examples, 1,199 val examples, and 5,794 testing examples. The task is to classify birds as waterbirds or landbirds while ignoring the background.

\textbf{Bird masks.} For every picture of the training set we use \detic{}~\cite{zhou2022detecting} to segment the bird in the image. The binary mask of the bird in the picture is denoted by $m$. We visually evaluate the quality of the masks and they are close to ideal. The most common error is that in rare cases there are additional birds in the background which are included in the mask.

\textbf{Pretrained representations.} We use three pretrained models in our experiments: ImageNet-pretrained ResNet-50,  SWAG-pretrained and ImageNet-finetuned RegNetY, and a self-supervised ViT-B/14 from DINOv2. These models produce $d$-dimensional representations for each, where $d=2048$, $d=7392$, $d=786$ for the three models respectively. \citet{mehta2022you} shows that the better models produce better worst group accuracy (WGA). 
Throughout this paper we use $z$ to denote representations.

\textbf{Attributing neurons.}
Whether or not individual neurons can be attributed to specific image regions is not clear.
There is some evidence against it and some evidence in support of it~\citep{elhage2022superposition}.
Certainly some layers and operations in deep learning give preference to the standard basis.
These include element-wise activations functions, batch normalization, dropout, etc.
We use \captum{}~\citep{kokhlikyan2020captum} to find this attribution. \captum{} has implementations of several attribution algorithms.
In our preliminary experiments, we found that the results with the Integrated Gradients method~\citep{sundararajan2017axiomatic} are good enough.

For a given image $x_j \in \bR^{224 \times 224}$, with foreground mask $m_j\in \mathset{0,1}^{224 \times 224}$, and representation $z_j \in \bR^d$, we use \captum{} to compute attribution heatmap $a^i_j \in \mathbb{R}^{224 \times 224}$ of the neuron $i$-th neuron over input pixels. We define spuriousness $s(x_j) \in \bR^d$ the following way:
\begin{equation}
    s_i(x_j) = \frac{\sum{(m_j \odot |a^i_j|)}}{\sum{m_j}} - \frac{\sum{((1-m_j) \odot |a^i_j|)}}{\sum{(1-m_j)}},
\end{equation}
where the sums are over pixels. The first term is the average attribution on the foreground pixels and the second term is the average attribution on the background pixels. Note that the attribution scores given by \captum{} can be both negative and positive, indicating the direction of the impact on the individual neuron. We use absolute values of the attribution scores as we are only interested in the magnitude of the impact. The spuriousness of the specific neuron $z_i$ is defined as the average spuriousness over the training set $X$:
\begin{equation}
    s_i = \frac{1}{|X|}\sum_{j=1}^{|X|}{s_i(x_j)}.
\end{equation}

\textbf{Linear models.} We train a linear classifier on a frozen representation using the \textit{scikit-learn} package. We use L-BFGS optimizer and disable regularization. We report overall accuracy and worst group accuracy on the test set, and mostly focus on optimizing the latter. We note that SGD-based methods to learn a linear classifier can discover drastically different solutions, including solutions with higher worst group accuracy. This is  especially true when early stopping is used. All linear models used in this paper use L-BFGS, and the analysis of SGD-based optimization is left for future work.

\textbf{Upper bound.} We compute an upper bound on worst group accuracy of L-BFGS-trained linear models by splitting the \emph{test set} into five equal parts of 1000 samples, and perform five-fold cross-validation. This way we ensure that each of the five models is trained on a subset of the same size as the training set, but with equal number of images with water backgrounds and land backgrounds. The average of worst group accuracies of the five models is reported as an upper bound.

\textbf{Statistical significance.} For selected experiments we performed bootstrapping on the training set to estimate the variability of the models with respect to changes in the training set. We resampled with repetitions five versions of the training set of the same size, and repeated the experiments on the five versions. We report the mean and standard deviation of the five metrics.

\section{Identifying Spurious Features}
We start our investigations by studying whether one can remove spurious features by removing individual neurons from the pretrained representations.

\begin{figure*}[t]
\begin{minipage}{0.33\textwidth}
    \begin{tikzpicture}[scale=0.7]
\begin{axis}[
    title={ResNet-50},
    xlabel={N},
    ylabel={},
    xmin=0, xmax=420,
    ymin=20, ymax=100,
    legend pos=south east,
    ymajorgrids=true,
    grid style=dashed,
    legend style={nodes={scale=0.65, transform shape}},
    legend columns=2,  
]

\addplot[
    color=red,
    mark=square,
    ]
    coordinates {
    (20,62.3)(40,65.0)(60,66.7)(80,67.8)(100,69.8)(120,65.0)(140,67.4)(160,68.2)(180,67.1)(200,67.4)(220,66.2)(240,65.3)(260,63.1)(280,58.6)(300,59.8)(320,61.2)(340,58.6)(360,59.0)(380,59.5)(400,59.8)
    };
    \addlegendentry{Captum}

    \addplot[
    color=blue,
    mark=square,
    ]
    coordinates {
    (20,25.7)(40,36.8)(60,43.5)(80,48.3)(100,50.2)(120,53.7)(140,54.0)(160,55.6)(180,57.3)(200,58.6)(220,58.7)(240,59.3)(260,61.2)(280,59.7)(300,57.0)(320,55.9)(340,58.3)(360,57.2)(380,57.5)(400,56.2)
    };
    \addlegendentry{Captum, Rot.}

\addplot[
    color=orange,
    mark=square,
    ]
    coordinates {
    (8,80.18018018018019)(16,82.70270270270271)(24,81.98198198198196)(32,81.98198198198199)(40,81.44144144144144)(48,81.08108108108108)(56,80.36036036036036)(64,82.52252252252254)(72,81.62162162162161)(80,82.34234234234232)(88,81.44144144144144)(96,81.44144144144143)(104,81.08108108108108)(112,82.70270270270271)(120,83.06306306306308)(128,82.34234234234235)
    };
    \addlegendentry{Captum,  GwAE[y]}

\addplot[
    color=green,
    mark=square,
    ]
    coordinates {
    (8,77.47747747747748)(16,83.78378378378379)(24,83.78378378378379)(32,79.27927927927928)(40,85.58558558558559)(48,85.58558558558559)(56,84.68468468468468)(64,83.78378378378379)(72,87.38738738738738)(80,83.78378378378379)(88,81.98198198198197)(96,81.08108108108108)(104,82.88288288288288)(112,84.68468468468468)(120,78.37837837837837)(128,81.08108108108108)
    };
    \addlegendentry{PCA,  GwAE[y]}



\addplot [orange, dashed] coordinates { (0,66.2) (420,66.2)};
\addlegendentry{N=2048 WGA (GwaE)}

\addplot [purple, dashed] coordinates { (0,78.7) (420,78.7)};
\addlegendentry{N=128 WGA (GwAE[y])}

\addplot [black, dashed] coordinates { (0,60.2) (420,60.2)};
\addlegendentry{N=2048 WGA (z)}

\end{axis}
\end{tikzpicture}
\end{minipage}%
\begin{minipage}{0.33\textwidth}
     \begin{tikzpicture}[scale=0.7]
\begin{axis}[
    title={RegNetY},
    xlabel={N},
    ylabel={},
    xmin=0, xmax=500,
    ymin=20, ymax=100,
    legend pos=south east,
    ymajorgrids=true,
    grid style=dashed,
    legend style={nodes={scale=0.65, transform shape}},
    legend columns=2,  
]

\addplot[
    color=red,
    mark=square,
    ]
    coordinates {
    (50,86.4)(100,85.7)(150,79.9)(200,85.4)(250,86.9)(300,86.9)(350,88.9)(400,88.8)(450,87.4)(500,87.2)(550,86.8)(600,86.3)(650,87.5)(700,86.3)(750,86.3)(800,86.4)(850,86.4)(900,86.9)(950,85.8)(1000,86.1)
    };
    \addlegendentry{Captum}

    \addplot[
    color=blue,
    mark=square,
    ]
    coordinates {
    (50,70.6)(100,74.6)(150,77.1)(200,84.6)(250,84.9)(300,84.1)(350,85.7)(400,86.6)(450,86.0)(500,85.4)(550,84.9)(600,83.8)(650,82.9)(700,82.4)(750,83.9)(800,83.6)(850,84.3)(900,83.6)(950,83.6)(1000,83.6)
    };
    \addlegendentry{Captum, Rot.}

\addplot[
    color=orange,
    mark=square,
    ]
    coordinates {
   (29,71.71171171171171)(58,80.54054054054053)(87,84.50450450450452)(116,85.04504504504504)(145,86.84684684684684)(174,87.74774774774775)(203,87.74774774774775)(232,88.64864864864866)(261,89.72972972972974)(290,89.1891891891892)(319,86.48648648648648)(348,89.54954954954954)(377,89.54954954954954)(406,90.990990990991)(435,91.17117117117118)(462,90.63063063063063)
    };
    \addlegendentry{Captum,  GwAE[y]}

\addplot[
    color=green,
    mark=square,
    ]
    coordinates {
    (29,91.8918918918919)(58,95.4954954954955)(87,90.09009009009009)(116,90.09009009009009)(145,95.4954954954955)(174,95.4954954954955)(203,86.48648648648648)(232,92.7927927927928)(261,92.7927927927928)(290,92.7927927927928)(319,91.8918918918919)(348,95.4954954954955)(377,92.7927927927928)(406,94.5945945945946)(435,93.69369369369369)(462,95.4954954954955)
    };
    \addlegendentry{PCA,  GwAE[y]}
    


\addplot [orange, dashed] coordinates { (0,83.0) (420,83.0)};
\addlegendentry{N=7392 WGA (GwAE)}

\addplot [purple, dashed] coordinates { (0,87.4) (1020,87.4)};
\addlegendentry{N=462 WGA (GwAE[y])}

\addplot [black, dashed] coordinates { (0,83.5) (420,83.5)};
\addlegendentry{N=7392 WGA (z)}
    
\end{axis}
\end{tikzpicture}
\end{minipage}
\begin{minipage}{0.33\textwidth}
    \begin{tikzpicture}[scale=0.7]
\begin{axis}[
    title={DINOv2},
    xlabel={N},
    ylabel={},
    xmin=0, xmax=420,
    ymin=20, ymax=100,
    legend pos=south west,
    ymajorgrids=true,
    grid style=dashed,
    legend style={nodes={scale=0.6, transform shape}},
    legend columns=2,  
]

\addplot[
    color=red,
    mark=square,
    ]
    coordinates {
    (20,58.6)(40,73.2)(60,72.4)(80,77.6)(100,80.1)(120,80.2)(140,81.2)(160,81.3)(180,81.5)(200,83.6)(220,82.9)(240,82.2)(260,84.4)(280,83.5)(300,82.4)(320,83.3)(340,82.9)(360,83.0)(380,83.6)(400,83.5)
    };
    \addlegendentry{Captum}

    \addplot[
    color=blue,
    mark=square,
    ]
    coordinates {
    (20,35.2)(40,56.1)(60,69.5)(80,70.2)(100,78.0)(120,81.3)(140,84.3)(160,79.8)(180,82.2)(200,84.4)(220,85.0)(240,84.1)(260,82.6)(280,84.9)(300,85.2)(320,86.3)(340,86.6)(360,85.0)(380,84.7)(400,85.2)
    };
    \addlegendentry{Captum, Rot.}

\addplot[
    color=orange,
    mark=square,
    ]
    coordinates {
    (6,3.6036036036036037)(12,32.792792792792795)(18,51.53153153153153)(24,60.00000000000001)(30,69.1891891891892)(36,70.09009009009009)(42,77.11711711711712)(48,78.91891891891892)(54,81.62162162162161)(60,81.98198198198199)(66,83.6036036036036)(72,82.52252252252254)(78,84.86486486486487)(84,85.94594594594595)(90,86.3063063063063)(96,87.02702702702702)
    };
    \addlegendentry{Captum,  GwAE[y]}

\addplot[
    color=green,
    mark=square,
    ]
    coordinates {
    (6,91.8918918918919)(12,95.4954954954955)(18,92.7927927927928)(24,93.69369369369369)(30,96.3963963963964)(36,95.4954954954955)(42,96.3963963963964)(48,94.5945945945946)(54,92.7927927927928)(60,94.5945945945946)(66,92.7927927927928)(72,92.7927927927928)(78,94.5945945945946)(84,91.8918918918919)(90,91.8918918918919)(96,90.990990990991)
    };
    \addlegendentry{PCA,  GwAE[y]}



\addplot [orange, dashed] coordinates { (0,88.9408) (420,88.9408)};
\addlegendentry{N=768 WGA (GwAE)}

\addplot [purple, dashed] coordinates { (0,93.5) (420,93.5)};
\addlegendentry{N=96 WGA (GwAE[y])}

\addplot [black, dashed] coordinates { (0,88.5) (420,88.5)};
\addlegendentry{N=768 WGA (z)}

\end{axis}
\end{tikzpicture}
\end{minipage}

\caption{Worst group accuracy of linear models on subsets of $N$ features selected using various methods.}
\label{fig:main}
\end{figure*}
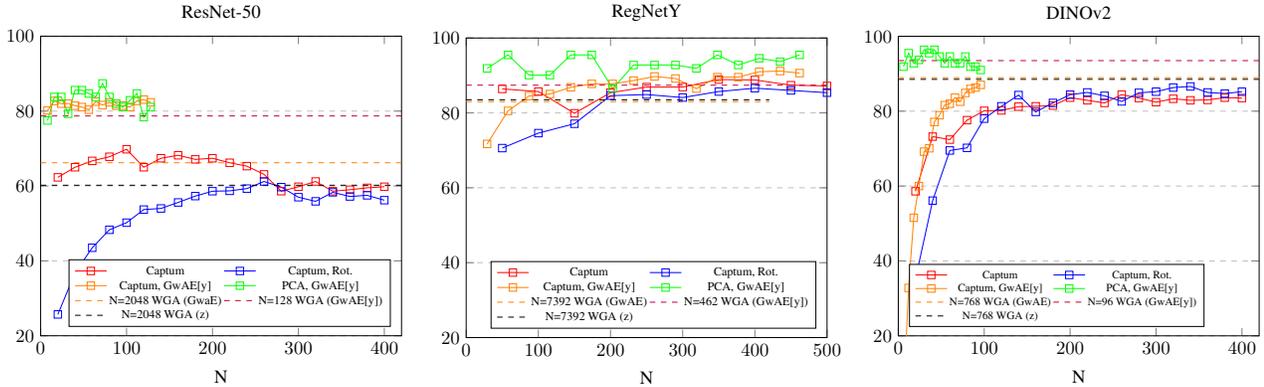

\textbf{Removing features that look mostly to background improves worst group accuracy.} We sort the $d$ neurons according to $s_i$ and consider keeping only the top $N$ neurons for the linear models. These models are denoted by $\text{Captum}^{N}(z, m)$.
\cref{fig:main} shows the results. 
In case of ResNet-50, a linear model on the top $N=50$ neurons significantly improves the baseline (see also \cref{tab:resnet}). 
The improvement is seen with up to $N=260$ neurons. 
This confirms the hypothesis that there are many spurious features that harm the worst group accuracy of linear models, and $s_i$ can be used as a measure to identify them.
A similar effect is observed with RegNetY, there is improvement for at least up to $N=1000$.

We could not detect this phenomenon in case of DINOv2. Keeping top neurons in terms of $s_i$ worsens the metrics. Most likely this means that the individual neurons are neither pure spurious nor pure non-spurious. A supporting evidence is that the variance of $s_i$ scores across the neurons is smaller than in case of ResNet-50.

\textbf{Pretrained representations are relatively axis-aligned.} To verify whether the directions in the representation space responsible for spurious features are aligned with the axes, we sample a random rotation matrix, apply it to the representations $z$, and recalculate $s_i$ for them. Then we pick the top neurons from the rotated space and train new linear models. These models are denoted by $\text{Captum}^{N}(\text{Rot}(z), m)$. As seen in \cref{fig:main}, the scores are significantly worse in ResNet-50 (for $N \leq 250$) and RegNetY (for $N \leq 150$). This implies that the spuriousness directions are aligned with axes for these two representations. The difference is much smaller in DINOv2, which is expected, as there were no distinctive spurious neurons  even before the rotation.

\section{Disentangling Spurious and Core Features}

\textbf{Designing a group-aware autoencoder.} As seen in the previous section, there exist spurious coordinates in ResNet-50 and RegNetY representations. We also showed that distinctive spuriousness is lost when we apply a random rotation matrix. This raises a question whether there exists another linear transformation that will make spurious features even more axis aligned, i.e. there will be new neurons that more specifically capture the spurious features. In other words, we are looking for ways to disentangle spurious and core features with a linear transformation.

We design a simple autoencoder where the linear encoder maps input $z$ to three vectors: $z_y$, $z_c$ and $z_n$. We force $z_y$ and $z_c$ to contain information about the label and the background, respectively. We do this by adding another linear layer on top of $z_y$ and $z_c$ that predict $\hat{y} = W_y z_y$ and $\hat{c} = W_c z_c$ which are supervised by the corresponding signals. The linear decoder takes the concatenation of $z_y$, $z_c$ and $z_n$ and reconstructs $\hat{z}$ which should be close to the original $z$. We assume that $z_n$ will store the rest of the information in $z$ that is not relevant for predicting either the label or the background. Note that we need group-level information to train this autoencoder, but we do not need the masks of the birds. Following \cite{discovery-separation-jaiswal}, we add an additional regularization term that minimizes mutual information between $z_y$ and $z_c$. The final loss function is the following:
\begin{equation}
    L = \text{ce}(\hat{y}, y) + \text{ce}(\hat{c}, c) + 10 \| \hat{z} - z \| + 50\ \text{HSIC}(z_y, z_c),
\end{equation}
where $\text{ce}(\cdot, \cdot)$ is the cross-entropy loss, and HSIC denotes the Hilbert-Schmidt Independence Criterion \cite{hsic}. 

We train the autoencoder on the \textit{upsampled} version of the training set so that each group is represented equally. This also justifies the minimization of the mutual information between $z_y$ and $z_c$, as they are correlated in the original training distribution. The sum of the dimensions of the three vectors $z_y$, $z_c$ and $z_n$ matches $d$ for each backbone. The linear model trained on top of the concatenation of the three vectors is denoted by $\text{GwAE}(z, g)$, where $\text{GwAE}$ denotes the linear encoder of the group-wise trained autoencoder, and $g$ refers to the group information required for the training. $\text{GwAE}_y(z, g)$ denotes only the $z_y$ part of the encoder's output. Note that the linear classifier trained on top of the autoencoder still belongs to the space of linear classifiers.

We expect the linear models trained on $z$ and on the full $\text{GwAE}(z, g)$ to perform similarly. We surprisingly see that this is not the case with ResNet-50. $\text{GwAE}(z, g)$ is better by 6 percentage points. One explanation is that the autoencoder is not ideal, some information is lost by the encoder, and luckily the lost information contains some of the spurious features. We leave a deeper analysis for the future work.

\textbf{Label-aware part has mostly good features.} For all backbones we see that $z_y$ gathers core features and the linear models trained on them have significantly better worst group accuracy. In case of ResNet-50 and DINOv2, the results are even better than the ones by \captum{}, which means that the autoencoder managed to isolate core features much better than it was possible by simply removing neurons in the original $z$. In case of RegNetY, we see that the linear model on $z_y$ performs as good as many models trained on \captum{}-filtered neurons. This means that RegNetY features were already disentangled. 

Furthermore, we apply \captum{} on top of the $z_y$ neurons to see whether we can still identify and remove spurious features in $z_y$ (the orange plot in \cref{fig:main}). This was successful only in case of ResNet-50.

\textbf{PCA helps.} To separate the core features, the linear encoder needs to shrink some directions (for example those corresponding to spurious features).
For this reason, one can hypothesize that most of the variance in $z_y$ will be along the core features.
This motivates applying principal component analysis (PCA) on $z_y$ to further remove non-core features.
Unlike \captum{}, PCA does not require additional information from the data.
We find that, indeed, training linear models on the $N$ principal components of $z_y$ still improves the worst group accuracy for ResNet and RegNetY.
We note that applying PCA directly on $z$ does not help to identify core features: the principal components of the original space usually contain spurious features.


\begin{table}[th]
\centering
\caption{Results with ResNet-50.}
\begin{tabular}{lcc}
\toprule
\textbf{Method} & \textbf{Accuracy} & \textbf{WGA} \\
\midrule
Standard training on $z$ & 83.1$\pm$1.2 & 60.2$\pm$1.8 \\
PCA$^{20}(z)$ & 81.7$\pm$0.8 & 51.3$\pm$2.0 \\
Captum$^{100}(z,m)$ & 88.0$\pm$0.3 & 67.7$\pm$1.3 \\
Captum$^{300}($Rot$(z),m)$ & 81.4$\pm$1.1 & 55.8$\pm$2.7 \\
GwAE$(z,g)[y]$ & 91.2$\pm$0.3 & 78.7$\pm$0.3 \\
PCA$^{inf}($GwAE$(z,g)[y])$ & 91.2$\pm$0.3 & 78.7$\pm$0.3 \\
PCA$^{20}($GwAE$(z,g)[y])$ & \best{93.9$\pm$0.2} & \best{81.7$\pm$1.1} \\
Captum$^{20}($GwAE$(z,g),m)$ & \best{93.5$\pm$0.1} & \best{81.0$\pm$0.6} \\
Captum$^{20}($GwAE$(z,g)[y],m)$ & \best{93.5$\pm$0.1} & \best{81.0$\pm$0.6} \\
\midrule
Upper bound on $z$ & 92.3$\pm$0.7 & 82.9$\pm$3.9 \\
Upper bound on GwAE$(z,g)$ & 92.6$\pm$1.1 & 81.1$\pm$5.0 \\
\bottomrule
\end{tabular}
\label{tab:resnet}
\end{table}

\section{Conclusion}
With carefully designed experiments we have shown that pretrained image representations contain spurious features that can be identified and removed to improve worst group accuracy of the linear models. In most representations it is also possible to disentangle spurious features and further improve the performance.
In future work we plan experiments with more backbones and datasets to see how well these findings generalize to other settings.
We also note that it is possible to find linear models with better worst group accuracy if we use stochastic gradient descent with early stopping. This analysis is also left for future work.


\bibliography{main}

\begin{thebibliography}{42}
\providecommand{\natexlab}[1]{#1}
\providecommand{\url}[1]{\texttt{#1}}
\expandafter\ifx\csname urlstyle\endcsname\relax
  \providecommand{\doi}[1]{doi: #1}\else
  \providecommand{\doi}{doi: \begingroup \urlstyle{rm}\Url}\fi

\bibitem[Arjovsky et~al.(2019)Arjovsky, Bottou, Gulrajani, and
  Lopez-Paz]{arjovsky2019invariant}
Arjovsky, M., Bottou, L., Gulrajani, I., and Lopez-Paz, D.
\newblock Invariant risk minimization.
\newblock \emph{arXiv preprint arXiv:1907.02893}, 2019.

\bibitem[Beery et~al.(2018)Beery, Van~Horn, and Perona]{beery2018recognition}
Beery, S., Van~Horn, G., and Perona, P.
\newblock Recognition in terra incognita.
\newblock In \emph{Proceedings of the European conference on computer vision
  (ECCV)}, pp.\  456--473, 2018.

\bibitem[Dosovitskiy et~al.(2021)Dosovitskiy, Beyer, Kolesnikov, Weissenborn,
  Zhai, Unterthiner, Dehghani, Minderer, Heigold, Gelly, Uszkoreit, and
  Houlsby]{dosovitskiy2021an}
Dosovitskiy, A., Beyer, L., Kolesnikov, A., Weissenborn, D., Zhai, X.,
  Unterthiner, T., Dehghani, M., Minderer, M., Heigold, G., Gelly, S.,
  Uszkoreit, J., and Houlsby, N.
\newblock An image is worth 16x16 words: Transformers for image recognition at
  scale.
\newblock In \emph{International Conference on Learning Representations}, 2021.

\bibitem[Elhage et~al.(2022)Elhage, Hume, Olsson, Schiefer, Henighan, Kravec,
  Hatfield-Dodds, Lasenby, Drain, Chen, Grosse, McCandlish, Kaplan, Amodei,
  Wattenberg, and Olah]{elhage2022superposition}
Elhage, N., Hume, T., Olsson, C., Schiefer, N., Henighan, T., Kravec, S.,
  Hatfield-Dodds, Z., Lasenby, R., Drain, D., Chen, C., Grosse, R., McCandlish,
  S., Kaplan, J., Amodei, D., Wattenberg, M., and Olah, C.
\newblock Toy models of superposition.
\newblock \emph{Transformer Circuits Thread}, 2022.

\bibitem[Galstyan et~al.(2022)Galstyan, Harutyunyan, Khachatrian, Steeg, and
  Galstyan]{galstyan2022failure}
Galstyan, T., Harutyunyan, H., Khachatrian, H., Steeg, G.~V., and Galstyan, A.
\newblock Failure modes of domain generalization algorithms.
\newblock In \emph{Proceedings of the IEEE/CVF Conference on Computer Vision
  and Pattern Recognition}, pp.\  19077--19086, 2022.

\bibitem[Ganin et~al.(2016)Ganin, Ustinova, Ajakan, Germain, Larochelle,
  Laviolette, Marchand, and Lempitsky]{ganin2016domain}
Ganin, Y., Ustinova, E., Ajakan, H., Germain, P., Larochelle, H., Laviolette,
  F., Marchand, M., and Lempitsky, V.
\newblock Domain-adversarial training of neural networks.
\newblock \emph{The journal of machine learning research}, 17\penalty0
  (1):\penalty0 2096--2030, 2016.

\bibitem[Geirhos et~al.(2019)Geirhos, Rubisch, Michaelis, Bethge, Wichmann, and
  Brendel]{geirhos2018imagenettrained}
Geirhos, R., Rubisch, P., Michaelis, C., Bethge, M., Wichmann, F.~A., and
  Brendel, W.
\newblock Imagenet-trained {CNN}s are biased towards texture; increasing shape
  bias improves accuracy and robustness.
\newblock In \emph{International Conference on Learning Representations}, 2019.

\bibitem[Geirhos et~al.(2020)Geirhos, Jacobsen, Michaelis, Zemel, Brendel,
  Bethge, and Wichmann]{geirhos2020shortcut}
Geirhos, R., Jacobsen, J.-H., Michaelis, C., Zemel, R., Brendel, W., Bethge,
  M., and Wichmann, F.~A.
\newblock Shortcut learning in deep neural networks.
\newblock \emph{Nature Machine Intelligence}, 2\penalty0 (11):\penalty0
  665--673, 2020.

\bibitem[Ghosal et~al.(2022)Ghosal, Ming, and Li]{ghosal2022vision}
Ghosal, S.~S., Ming, Y., and Li, Y.
\newblock Are vision transformers robust to spurious correlations?
\newblock \emph{arXiv preprint arXiv:2203.09125}, 2022.

\bibitem[Gretton et~al.(2005)Gretton, Bousquet, Smola, and Sch{\"o}lkopf]{hsic}
Gretton, A., Bousquet, O., Smola, A., and Sch{\"o}lkopf, B.
\newblock Measuring statistical dependence with hilbert-schmidt norms.
\newblock In \emph{Algorithmic Learning Theory: 16th International Conference,
  ALT 2005, Singapore, October 8-11, 2005. Proceedings 16}, pp.\  63--77.
  Springer, 2005.

\bibitem[Gulrajani \& Lopez-Paz(2021)Gulrajani and Lopez-Paz]{gulrajani2021in}
Gulrajani, I. and Lopez-Paz, D.
\newblock In search of lost domain generalization.
\newblock In \emph{International Conference on Learning Representations}, 2021.

\bibitem[Gururangan et~al.(2018)Gururangan, Swayamdipta, Levy, Schwartz,
  Bowman, and Smith]{gururangan-etal-2018-annotation}
Gururangan, S., Swayamdipta, S., Levy, O., Schwartz, R., Bowman, S., and Smith,
  N.~A.
\newblock Annotation artifacts in natural language inference data.
\newblock In \emph{Proceedings of the 2018 Conference of the North {A}merican
  Chapter of the Association for Computational Linguistics: Human Language
  Technologies, Volume 2 (Short Papers)}, pp.\  107--112, New Orleans,
  Louisiana, June 2018. Association for Computational Linguistics.
\newblock \doi{10.18653/v1/N18-2017}.

\bibitem[Hermann \& Lampinen(2020)Hermann and Lampinen]{hermann2020shapes}
Hermann, K. and Lampinen, A.
\newblock What shapes feature representations? exploring datasets,
  architectures, and training.
\newblock \emph{Advances in Neural Information Processing Systems},
  33:\penalty0 9995--10006, 2020.

\bibitem[Izmailov et~al.(2022)Izmailov, Kirichenko, Gruver, and
  Wilson]{izmailov2022feature}
Izmailov, P., Kirichenko, P., Gruver, N., and Wilson, A.~G.
\newblock On feature learning in the presence of spurious correlations.
\newblock \emph{Advances in Neural Information Processing Systems},
  35:\penalty0 38516--38532, 2022.

\bibitem[Jaiswal et~al.(2019)Jaiswal, Brekelmans, Moyer, Steeg, AbdAlmageed,
  and Natarajan]{discovery-separation-jaiswal}
Jaiswal, A., Brekelmans, R., Moyer, D., Steeg, G.~V., AbdAlmageed, W., and
  Natarajan, P.
\newblock Discovery and separation of features for invariant representation
  learning.
\newblock \emph{arXiv preprint arXiv:1912.00646}, 2019.

\bibitem[Kirichenko et~al.(2023)Kirichenko, Izmailov, and
  Wilson]{kirichenko2023last}
Kirichenko, P., Izmailov, P., and Wilson, A.~G.
\newblock Last layer re-training is sufficient for robustness to spurious
  correlations.
\newblock In \emph{The Eleventh International Conference on Learning
  Representations}, 2023.

\bibitem[Koh et~al.(2021)Koh, Sagawa, Marklund, Xie, Zhang, Balsubramani, Hu,
  Yasunaga, Phillips, Gao, et~al.]{koh2021wilds}
Koh, P.~W., Sagawa, S., Marklund, H., Xie, S.~M., Zhang, M., Balsubramani, A.,
  Hu, W., Yasunaga, M., Phillips, R.~L., Gao, I., et~al.
\newblock Wilds: A benchmark of in-the-wild distribution shifts.
\newblock In \emph{International Conference on Machine Learning}, pp.\
  5637--5664. PMLR, 2021.

\bibitem[Kokhlikyan et~al.(2020)Kokhlikyan, Miglani, Martin, Wang, Alsallakh,
  Reynolds, Melnikov, Kliushkina, Araya, Yan, and
  Reblitz-Richardson]{kokhlikyan2020captum}
Kokhlikyan, N., Miglani, V., Martin, M., Wang, E., Alsallakh, B., Reynolds, J.,
  Melnikov, A., Kliushkina, N., Araya, C., Yan, S., and Reblitz-Richardson, O.
\newblock Captum: A unified and generic model interpretability library for
  pytorch, 2020.

\bibitem[Li et~al.(2018)Li, Tian, Gong, Liu, Liu, Zhang, and Tao]{li2018deep}
Li, Y., Tian, X., Gong, M., Liu, Y., Liu, T., Zhang, K., and Tao, D.
\newblock Deep domain generalization via conditional invariant adversarial
  networks.
\newblock In \emph{Proceedings of the European conference on computer vision
  (ECCV)}, pp.\  624--639, 2018.

\bibitem[Lu et~al.(2022)Lu, Wu, Hern{\'a}ndez-Lobato, and
  Sch{\"o}lkopf]{lu2022invariant}
Lu, C., Wu, Y., Hern{\'a}ndez-Lobato, J.~M., and Sch{\"o}lkopf, B.
\newblock Invariant causal representation learning for out-of-distribution
  generalization.
\newblock In \emph{International Conference on Learning Representations}, 2022.

\bibitem[McCoy et~al.(2019)McCoy, Pavlick, and Linzen]{mccoy-etal-2019-right}
McCoy, T., Pavlick, E., and Linzen, T.
\newblock Right for the wrong reasons: Diagnosing syntactic heuristics in
  natural language inference.
\newblock In \emph{Proceedings of the 57th Annual Meeting of the Association
  for Computational Linguistics}, pp.\  3428--3448, Florence, Italy, July 2019.
  Association for Computational Linguistics.
\newblock \doi{10.18653/v1/P19-1334}.

\bibitem[Mehta et~al.(2022)Mehta, Albiero, Chen, Evtimov, Glaser, Li, and
  Hassner]{mehta2022you}
Mehta, R., Albiero, V., Chen, L., Evtimov, I., Glaser, T., Li, Z., and Hassner,
  T.
\newblock You only need a good embeddings extractor to fix spurious
  correlations.
\newblock \emph{arXiv preprint arXiv:2212.06254}, 2022.

\bibitem[Menon et~al.(2021)Menon, Rawat, and
  Kumar]{menon2021overparameterisation}
Menon, A.~K., Rawat, A.~S., and Kumar, S.
\newblock Overparameterisation and worst-case generalisation: friend or foe?
\newblock In \emph{International Conference on Learning Representations}, 2021.

\bibitem[Muandet et~al.(2013)Muandet, Balduzzi, and
  Sch{\"o}lkopf]{muandet2013domain}
Muandet, K., Balduzzi, D., and Sch{\"o}lkopf, B.
\newblock Domain generalization via invariant feature representation.
\newblock In \emph{International conference on machine learning}, pp.\  10--18.
  PMLR, 2013.

\bibitem[Nam et~al.(2020)Nam, Cha, Ahn, Lee, and Shin]{nam2020learning}
Nam, J., Cha, H., Ahn, S., Lee, J., and Shin, J.
\newblock Learning from failure: De-biasing classifier from biased classifier.
\newblock \emph{Advances in Neural Information Processing Systems},
  33:\penalty0 20673--20684, 2020.

\bibitem[Pezeshki et~al.(2021)Pezeshki, Kaba, Bengio, Courville, Precup, and
  Lajoie]{pezeshki2021gradient}
Pezeshki, M., Kaba, O., Bengio, Y., Courville, A.~C., Precup, D., and Lajoie,
  G.
\newblock Gradient starvation: A learning proclivity in neural networks.
\newblock \emph{Advances in Neural Information Processing Systems},
  34:\penalty0 1256--1272, 2021.

\bibitem[Ribeiro et~al.(2016)Ribeiro, Singh, and Guestrin]{ribeiro2016should}
Ribeiro, M.~T., Singh, S., and Guestrin, C.
\newblock " why should i trust you?" explaining the predictions of any
  classifier.
\newblock In \emph{Proceedings of the 22nd ACM SIGKDD international conference
  on knowledge discovery and data mining}, pp.\  1135--1144, 2016.

\bibitem[Sagawa* et~al.(2020)Sagawa*, Koh*, Hashimoto, and
  Liang]{Sagawa*2020Distributionally}
Sagawa*, S., Koh*, P.~W., Hashimoto, T.~B., and Liang, P.
\newblock Distributionally robust neural networks.
\newblock In \emph{International Conference on Learning Representations}, 2020.

\bibitem[Sagawa et~al.(2020)Sagawa, Koh, Hashimoto, and Liang]{group_dro}
Sagawa, S., Koh, P.~W., Hashimoto, T.~B., and Liang, P.
\newblock Distributionally robust neural networks for group shifts: On the
  importance of regularization for worst-case generalization, 2020.

\bibitem[Shah et~al.(2020)Shah, Tamuly, Raghunathan, Jain, and
  Netrapalli]{shah2020pitfalls}
Shah, H., Tamuly, K., Raghunathan, A., Jain, P., and Netrapalli, P.
\newblock The pitfalls of simplicity bias in neural networks.
\newblock \emph{Advances in Neural Information Processing Systems},
  33:\penalty0 9573--9585, 2020.

\bibitem[Shi et~al.(2023)Shi, Daunhawer, Vogt, Torr, and Sanyal]{shi2023robust}
Shi, Y., Daunhawer, I., Vogt, J.~E., Torr, P., and Sanyal, A.
\newblock How robust is unsupervised representation learning to distribution
  shift?
\newblock In \emph{The Eleventh International Conference on Learning
  Representations}, 2023.

\bibitem[Sun \& Saenko(2016)Sun and Saenko]{sun2016deep}
Sun, B. and Saenko, K.
\newblock Deep coral: Correlation alignment for deep domain adaptation.
\newblock In \emph{Computer Vision--ECCV 2016 Workshops: Amsterdam, The
  Netherlands, October 8-10 and 15-16, 2016, Proceedings, Part III 14}, pp.\
  443--450. Springer, 2016.

\bibitem[Sundararajan et~al.(2017)Sundararajan, Taly, and
  Yan]{sundararajan2017axiomatic}
Sundararajan, M., Taly, A., and Yan, Q.
\newblock Axiomatic attribution for deep networks.
\newblock In \emph{International conference on machine learning}, pp.\
  3319--3328. PMLR, 2017.

\bibitem[Torralba \& Efros(2011)Torralba and Efros]{torralba2011unbiased}
Torralba, A. and Efros, A.~A.
\newblock Unbiased look at dataset bias.
\newblock In \emph{CVPR 2011}, pp.\  1521--1528. IEEE, 2011.

\bibitem[Wang et~al.(2019{\natexlab{a}})Wang, Ge, Lipton, and
  Xing]{wang2019learning}
Wang, H., Ge, S., Lipton, Z., and Xing, E.~P.
\newblock Learning robust global representations by penalizing local predictive
  power.
\newblock \emph{Advances in Neural Information Processing Systems}, 32,
  2019{\natexlab{a}}.

\bibitem[Wang et~al.(2019{\natexlab{b}})Wang, He, and Xing]{wang2018learning}
Wang, H., He, Z., and Xing, E.~P.
\newblock Learning robust representations by projecting superficial statistics
  out.
\newblock In \emph{International Conference on Learning Representations},
  2019{\natexlab{b}}.

\bibitem[Welinder et~al.(2010)Welinder, Branson, Mita, Wah, Schroff, Belongie,
  and Perona]{cub_dataset}
Welinder, P., Branson, S., Mita, T., Wah, C., Schroff, F., Belongie, S., and
  Perona, P.
\newblock {Caltech-UCSD Birds 200}.
\newblock Technical Report CNS-TR-2010-001, California Institute of Technology,
  2010.

\bibitem[Xiao et~al.(2021)Xiao, Engstrom, Ilyas, and Madry]{xiao2021noise}
Xiao, K.~Y., Engstrom, L., Ilyas, A., and Madry, A.
\newblock Noise or signal: The role of image backgrounds in object recognition.
\newblock In \emph{International Conference on Learning Representations}, 2021.

\bibitem[Zech et~al.(2018)Zech, Badgeley, Liu, Costa, Titano, and
  Oermann]{zech2018variable}
Zech, J.~R., Badgeley, M.~A., Liu, M., Costa, A.~B., Titano, J.~J., and
  Oermann, E.~K.
\newblock Variable generalization performance of a deep learning model to
  detect pneumonia in chest radiographs: a cross-sectional study.
\newblock \emph{PLoS medicine}, 15\penalty0 (11):\penalty0 e1002683, 2018.

\bibitem[Zhao et~al.(2020)Zhao, Gong, Liu, Fu, and Tao]{zhao2020domain}
Zhao, S., Gong, M., Liu, T., Fu, H., and Tao, D.
\newblock Domain generalization via entropy regularization.
\newblock \emph{Advances in Neural Information Processing Systems},
  33:\penalty0 16096--16107, 2020.

\bibitem[Zhou et~al.(2018)Zhou, Lapedriza, Khosla, Oliva, and
  Torralba]{places_dataset}
Zhou, B., Lapedriza, {\`A}., Khosla, A., Oliva, A., and Torralba, A.
\newblock Places: A 10 million image database for scene recognition.
\newblock \emph{IEEE Transactions on Pattern Analysis and Machine
  Intelligence}, 40:\penalty0 1452--1464, 2018.

\bibitem[Zhou et~al.(2022)Zhou, Girdhar, Joulin, Kr{\"a}henb{\"u}hl, and
  Misra]{zhou2022detecting}
Zhou, X., Girdhar, R., Joulin, A., Kr{\"a}henb{\"u}hl, P., and Misra, I.
\newblock Detecting twenty-thousand classes using image-level supervision.
\newblock In \emph{Computer Vision--ECCV 2022: 17th European Conference, Tel
  Aviv, Israel, October 23--27, 2022, Proceedings, Part IX}, pp.\  350--368.
  Springer, 2022.

\end{thebibliography}
\bibliographystyle{icml2023}

\newpage
\appendix
\onecolumn
\section{Appendix}

In \cref{tab:dino,tab:regnety} we show the detailed results for RegNetY and DINOv2.

\begin{table}[h]
\caption{Results with RegNetY.}
\label{tab:regnety}

\centering
\begin{tabular}{lcc}
\toprule
\textbf{} & \textbf{Accuracy} & \textbf{WGA} \\
\midrule
$z$ & 95.2$\pm$0.3 & 83.5$\pm$2.2 \\
PCA$^{20}(z)$ & 90.7$\pm$0.5 & 74.6$\pm$1.7 \\
Captum$^{350}(z,m)$ & 94.9$\pm$0.3 & 88.0$\pm$1.7 \\
Captum$^{350}($Rot$(z),m)$ & 93.5$\pm$0.8 & 85.6$\pm$1.1 \\
GwAE$(z,g)[y]$ & \best{96.4$\pm$0.3} & 87.4$\pm$0.9 \\
PCA$^{inf}($GwAE$(z,g)[y])$ & 95.9$\pm$0.3 & \best{89.7$\pm$0.9} \\
PCA$^{20}($GwAE$(z,g)[y])$ & \best{96.5$\pm$0.1} & \best{90.4$\pm$1.7} \\
Captum$^{20}($GwAE$(z,g),m)$ & \best{96.3$\pm$0.2} & 88.9$\pm$0.3 \\
Captum$^{20}($GwAE$(z,g)[y],m)$ & \best{96.3$\pm$0.2} & 88.9$\pm$0.3 \\
\midrule
Upper bound on $z$ & 98.4$\pm$0.3 & 94.1$\pm$2.0 \\
Upper bound on GwAE$(z,g)$ & 98.3$\pm$0.3 & 92.8$\pm$2.3 \\ \hline
\end{tabular}
\end{table}

\begin{table}[h]
\centering
\caption{Results with DINOv2.}
\label{tab:dino}
\begin{tabular}{lcc}
\toprule
\textbf{} & \textbf{Accuracy} & \textbf{WGA} \\
\midrule
$z$ & 95.9$\pm$0.3 & 88.5$\pm$0.9 \\
PCA$^{20}(z)$ & 93.7$\pm$0.4 & 80.6$\pm$1.4 \\
Captum$^{100}(z,m)$ & 92.1$\pm$0.6 & 79.8$\pm$0.6 \\
Captum$^{100}($Rot$(z),m)$ & 92.1$\pm$0.1 & 76.6$\pm$2.4 \\
Captum$^{700}(z,m)$ & 96.3$\pm$0.3 & 88.6$\pm$0.6 \\
Captum$^{700}($Rot$(z),m)$ & 96.3$\pm$0.4 & 89.2$\pm$1.9 \\
GwAE$(z,g)[y]$ & 96.6$\pm$0.3 & 93.5$\pm$0.4 \\
PCA$^{inf}($GwAE$(z,g)[y])$ & 96.8$\pm$0.2 & 93.0$\pm$0.6 \\
PCA$^{20}($GwAE$(z,g)[y])$ & \best{97.4$\pm$0.2} & \best{94.0$\pm$0.8} \\
Captum$^{100}($GwAE$(z,g),m)$ & 96.6$\pm$0.4 & 90.6$\pm$0.8 \\
Captum$^{20}($GwAE$(z,g),m)$ & 94.1$\pm$0.2 & 83.8$\pm$1.2 \\
Captum$^{20}($GwAE$(z,g)[y],m)$ & 97.0$\pm$0.1 & 92.7$\pm$0.5 \\
\midrule
Upper bound on $z$ & 98.3$\pm$0.1 & 94.6$\pm$1.1 \\
Upper bound on GwAE$(z,g)$ & 98.0$\pm$0.2 & 93.9$\pm$1.9 \\ \hline
\end{tabular}
\end{table}

\end{document}